\documentclass[a4paper]{article}

\usepackage{INTERSPEECH2019}
\usepackage{xcolor}
\usepackage{enumerate}
\usepackage{diagbox}
\usepackage{enumitem}
\usepackage{multirow}

\title{From Semi-supervised to Almost-unsupervised Speech Recognition with Very-low Resource by Jointly Learning Phonetic Structures from Audio and Text Embeddings}
\name{Yi-Chen Chen, Sung-Feng Huang, Hung-yi Lee, Lin-shan Lee}
\address{National Taiwan University, Taiwan}
\email{piercing.zs0half@gmail.com,r06942045@ntu.edu.tw,\\tlkagkb93901106@gmail.com,lslee@gate.sinica.edu.tw}

\begin{document}

\setlength{\abovedisplayskip}{1pt}
\setlength{\belowdisplayskip}{1pt}
\setlength{\abovedisplayshortskip}{1pt}
\setlength{\belowdisplayshortskip}{1pt}

\maketitle
%

\begin{abstract}
\vspace{-1mm}
Producing a large amount of annotated speech data for training ASR systems remains difficult for more than 95\% of languages all over the world which are low-resourced.
However, we note human babies start to learn the language by the sounds (or phonetic structures) of a small number of exemplar words, and ``generalize" such knowledge to other words without hearing a large amount of data.
We initiate some preliminary work in this direction.
Audio Word2Vec is used to learn the phonetic structures from spoken words (signal segments), while another autoencoder is used to learn the phonetic structures from text words.
The relationships among the above two can be learned jointly, or separately after the above two are well trained.
This relationship can be used in speech recognition with very low resource.
In the initial experiments on the TIMIT dataset, only 2.1 hours of speech data (in which 2500 spoken words were annotated and the rest unlabeled) gave a word error rate of 44.6\%, and this number can be reduced to 34.2\% if 4.1 hr of speech data (in which 20000 spoken words were annotated) were given.
These results are not satisfactory, but a good starting point.
\end{abstract}
\vspace{-1mm}

\section{Introduction}
\label{sec:intro}
\vspace{-1mm}

Automatic speech recognition (ASR) has achieved remarkable success in many applications~\cite{bahdanau2016end,amodei2016deep,zhang2017very}.
However, with existing technologies, machines have to learn from a huge amount of annotated data to achieve acceptable accuracy, which makes the development of such technologies for new languages with low resource challenging. 
Collecting a large amount of speech data is expensive, not to mention having the data annotated.
This remains true for at least 95\% of languages all over the world.

Substantial effort has been reported on semi-supervised ASR~\cite{vesely2013semi,dikici2016semi,thomas2013deep,grezl2014combination,vesely2017semi,chen2018almost,karita2018semi,drexler2018combining}.
However, in most cases a large amount of speech data including a good portion annotated were still needed.
So training ASR systems with relatively little data, most of which are not annotated, remains to be an important but unsolved problem.
Speech recognition under such ``very-low" resource conditions is the target task of this paper.

We note human babies start to learn the language by the sounds of a small number of exemplar words without hearing a large amount of data.
They more or less learn those words by ``how they sound", or the phonetic structures for the words.
These exemplar words and their phonetic structures then seem to ``generalize" to other words and sentences they learn later on.
It is certainly highly desired if machines can do that too.
In this paper we initiate some preliminary work in this direction.

Audio Word2Vec was proposed to transform spoken words (signal segments for words without knowing the underlying words they represent) to vectors of fixed dimensionality~\cite{DBLP:conf/interspeech/ChungWSLL16} carrying information about the phonetic structures of the spoken words.
Segmental Audio Word2Vec was further proposed to jointly segment an utterance into a sequence of spoken words and transform them into a sequence of vectors~\cite{SSAE}.
Substantial effort has been made to try to align such audio embeddings with word embeddings~\cite{chung2018unsupervised}, which was one way to teach machines to learn the words jointly with their sounds or phonetic structures.
Approaches of semi-supervised end-to-end speech recognition approaches along similar directions were also reported recently~\cite{karita2018semi,drexler2018combining}.
But all these works still used relatively large amount of training data.
On the other hand, unsupervised phoneme recognition and almost-unsupervised word recognition were recently achieved to some extent using zero or close-to-zero aligned audio and text data~\cite{DBLP:conf/interspeech/LiuCLL18,chen2018almost}, primarily by mapping the audio embeddings with text tokens, whose ``very-low" resource setting is the goal of this paper.

In this work, we let the machines learn the phonetic structures of words from the embedding spaces of respective spoken and text words, as well as the relationships between the two.
All these can be learned jointly, or separately for spoken and text words individually followed by learning the relationships between the two.
It was found the former is better, and reasonable speech recognition was achievable with very low resource.
In the initial experiments on the TIMIT dataset, only 2.1 hours of total speech data (in which 2500 spoken words were annotated and the rest unlabeled) gave a word error rate of 44.6\%, and this number can be reduced to 34.2\% if 4.1 hr of speech data (in which 20000 spoken words were annotated) were given.
These results are not satisfactory, but a good starting point.
\vspace{-1mm}

\begin{figure}[t]
  \centering
  \includegraphics[width=\linewidth]{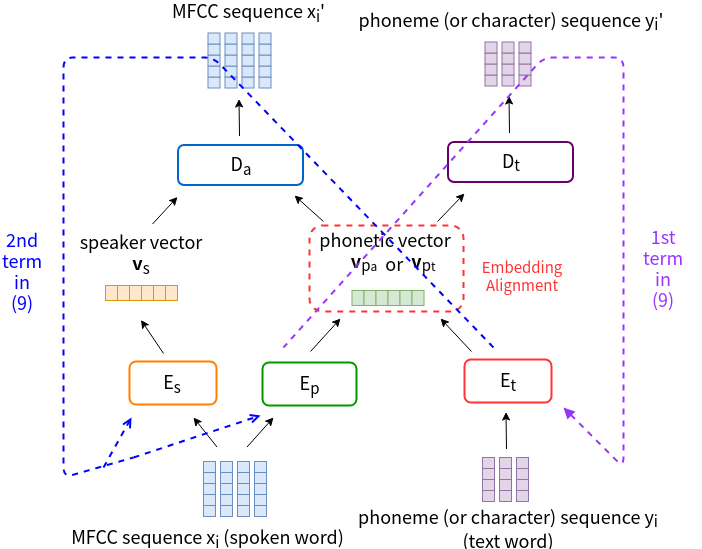}
  \caption{The architecture of the proposed approach.}
  \label{fig:overview}
\vspace{-6mm}
\end{figure}

\begin{figure}[t]
  \centering
  \includegraphics[width=\linewidth]{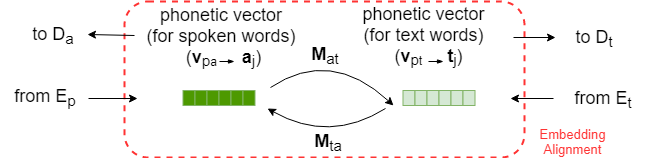}
  \caption{Embedding alignment (red dotted block in middle of Figure~\ref{fig:overview}) realized by transformation between two latent spaces.}
  \label{fig:alignment}
\vspace{-5mm}
\end{figure}

\section{Proposed Approach}
\label{sec:proposed}
\vspace{-1mm}

For clarity, we denote the speech corpus as $\mathbf{X} = {\{\mathbf{x}_{i}\}}_{i=1}^{M}$, which consists of $M$ spoken words, each represented as $\mathbf{x}_i=(\mathbf{x}_{i_1}, \mathbf{x}_{i_2}, ..., \mathbf{x}_{i_T})$, where $\mathbf{x}_{i_t}$ is the acoustic feature vector at time t and $T$ is the length of the spoken word. 
Each spoken word $\mathbf{x}_i$ corresponds to a text word in $\mathbf{W} = {\{w_{k}\}}_{k=1}^{N}$, where $N$ is the number of distinct text words in the corpus. 
We can represent each text word as a sequence of subword units, like phonemes or characters, and denote it as $\mathbf{y}_i=(\mathbf{y}_{i_1}, \mathbf{y}_{i_2}, ..., \mathbf{y}_{i_L})$, where $\mathbf{y}_{i_l}$ is the one-hot vector for the $l$\textsuperscript{th} subword and $L$ is the number of subwords in the word.
A small set of known paired data is denoted as $\mathbf{Z} = {\{(\mathbf{x}_{j}, \mathbf{y}_{j})\}}$, where $(\mathbf{x}_{j}, \mathbf{y}_{j})$ corresponds to the same text word.


In the initial work here we focus on the joint learning of words in audio and text forms, so we assume all training spoken words have been properly segmented with good boundaries.
Many existing approaches can be used to segment utterances into spoken words automatically~\cite{DBLP:conf/naacl/TranTBGLO18,DBLP:journals/jstsp/TangLKGLDSR17,kamper2017embedded,kamper2017segmental,SRAILICASSP15,WordEmbedIS14,QbyELSTMICASSP15,settle2017query,SemanticRepresentationICASSP18}, including the Segmental Audio Word2Vec~\cite{SSAE} mentioned above.
Extension to entire utterance input without segmentation is left for future work.

A text word corresponds to many different spoken words with varying acoustic factors such as speaker or microphone characteristics, and noise.
We jointly refer to all such acoustic factors as speaker characteristics below for simplicity.
\vspace{-1mm}

\subsection{Intra-domain Unsupervised Autoencoder Architecture}
\label{subsec:unsupervised}

\vspace{-1mm}
There are three encoders and two decoders in the architecture of the proposed approach in Figure~\ref{fig:overview}.
We use two encoders $E_p$ and $E_s$ to encode the phonetic structures and speaker characteristics of a spoken word $\mathbf{x}_i$ into an audio phonetic vector $\mathbf{v_{p_a}}$ and a speaker vector $\mathbf{v_s}$ respectively.
Meanwhile, we use another encoder $E_t$ to encode the phonetic structure of a text word $\mathbf{y}_i$ into a text phonetic vector $\mathbf{v_{p_t}}$, where text words $\mathbf{y}_i$ are represented as sequences of one-hot vectors for subwords. 

The audio decoder $D_a$ takes the pair ($\mathbf{v_{p_a}}$, $\mathbf{v_s}$) as input and reconstruct the original spoken word features $\mathbf{x}_i'$.
The text decoder $D_t$ takes $\mathbf{v_{p_t}}$ as input and reconstruct the original text word features $\mathbf{y}_i'$.
Two intra-domain losses are used for unsupervised training:
\begin{enumerate}[label=\arabic*), wide, labelwidth=!, labelindent=0pt]
    \item Intra-domain audio reconstruction loss, which is the mean-square-error between the audio original and the reconstructed features: \begin{equation}
    \begin{aligned} \label{in_a_r_loss}
    L_{in\_a\_r} &= \sum_{i} \| \mathbf{x}_i - D_a(E_p(\mathbf{x}_i), E_s(\mathbf{x}_i)) \|_2^2. 
    \end{aligned}
    \end{equation} 
    \item Intra-domain text reconstruction loss, which is the negative log-likelihood for the text vector sequences to be reconstructed: \begin{equation}
    \begin{aligned} \label{in_t_r_loss}
    L_{in\_t\_r} &= - \sum_{i} logPr(\mathbf{y}_i | D_t(E_t(\mathbf{y}_i))).
    \end{aligned}
    \end{equation} 
\end{enumerate}


\vspace{-4mm}

\subsection{Cross-domain Reconstruction with Paired Data}
\label{subsec:cross-domain}

\vspace{-1mm}
When the latent spaces for the phonetic structures for spoken words $\mathbf{x}_i$ and text words $\mathbf{y}_i$ are individually learned based on the intra-domain reconstruction losses (\ref{in_a_r_loss})(\ref{in_t_r_loss}), they can be very different, since the former are continuous signals with varying length and behavior, while the latter are sequences of discrete symbols with given length.
So here we try to learn them jointly using relatively small number of known pairs of spoken words $\mathbf{x}_j$ and the corresponding text words $\mathbf{y}_j$, $\mathbf{Z} = {\{(\mathbf{x}_{j}, \mathbf{y}_{j})\}}$. Hopefully the two latent spaces can be twisted somehow and end up with a single common latent space, in which both phonetic vectors for audio and text, $\mathbf{v_{p_a}}$ and $\mathbf{v_{p_t}}$, can be properly represented.
So two cross-domain losses below are used:
\begin{enumerate}[label=\arabic*), wide, labelwidth=!, labelindent=0pt]
    \setcounter{enumi}{2}
    \item Cross-domain audio reconstruction loss: \begin{equation}
    \begin{aligned}
    L_{cr\_a\_r} &= \sum_{(\mathbf{x}_j, \mathbf{y}_j) \in \mathbf{Z}} \| \mathbf{x}_j - D_a(E_t(\mathbf{y}_j), E_s(\mathbf{x}_j)) \|_2^2. 
    \label{cr_a_r_loss}
    \end{aligned}
    \end{equation} 
    \item Cross-domain text reconstruction loss: \begin{equation}
    \begin{aligned}
    L_{cr\_t\_r} &= - \sum_{(\mathbf{x}_j, \mathbf{y}_j) \in \mathbf{Z}} logPr(\mathbf{y}_j | D_t(E_p(\mathbf{x}_j))).
    \label{cr_t_r_loss}
    \end{aligned}
    \end{equation}
\end{enumerate}
By minimizing the reconstruction loss for the audio/text features obtained with the phonetic vectors computed from input sequences in the other domain as in (\ref{cr_a_r_loss})(\ref{cr_t_r_loss}), the phonetic vectors of spoken and text words can be somehow aligned to carry some consistent information about the phonetic structures.
\vspace{-1mm}

\subsection{Cross-domain Alignment of Phonetic Vectors with Paired Data}
\label{subsec:emb_align}

\vspace{-1mm}
On top of the cross-domain reconstruction losses (\ref{cr_a_r_loss})(\ref{cr_t_r_loss}), the two latent spaces can be further aligned by a hinge loss for all known pairs of spoken and text words $(\mathbf{x}_j, \mathbf{y}_j)$:
\begin{enumerate}[label=\arabic*), wide, labelwidth=!, labelindent=0pt]
    \setcounter{enumi}{4}
    \item Cross-domain embedding loss: \begin{equation}
    \begin{aligned}
    L_{cr\_emb} &= \sum_{(\mathbf{x}_j, \mathbf{y}_j) \in \mathbf{Z}} \| E_p(\mathbf{x}_j) - E_t(\mathbf{y}_j) \|_2^2 \\ &+ \sum_{(\mathbf{x}_i, \mathbf{y}_j) \notin \mathbf{Z}} \max(0, \lambda - \| E_p(\mathbf{x}_i) - E_t(\mathbf{y}_j) \|_2^2).
    \label{cr_emb_loss}
    \end{aligned}
    \end{equation} 
\end{enumerate}
In the second term of (\ref{cr_emb_loss}), for each text word $\mathbf{y}_j$, we randomly sample $\mathbf{x}_i$ such that $(\mathbf{x}_i, \mathbf{y}_j) \notin \mathbf{Z}$ to serve as a negative sample.
In this way, the phonetic vectors corresponding to different text words can be kept far enough apart.
Here in (\ref{cr_a_r_loss})(\ref{cr_t_r_loss})(\ref{cr_emb_loss}) the small number of paired spoken and text words $\{(\mathbf{x}_j$, $\mathbf{y}_j)\} \in \mathbf{Z}$ serve just as the small number of exemplar words and their sounds when human babies start to learn the language.
The reconstruction and alignment across the two spaces is somehow to try to ``generalize" the phonetic structures of these exemplar words to other words in the language as human babies do.
\vspace{-1mm}

\subsection{Joint Learning and Inference}
\label{subsec:inference}

\vspace{-1mm}
The total loss function $L$ to be minimized during training is the weighted sum of the above five losses:
\begin{equation}
\begin{aligned}
L &= \alpha_1 L_{in\_a\_r} + \alpha_2 L_{in\_t\_r} \\ 
&+ \alpha_3 L_{cr\_a\_r} + \alpha_4 L_{cr\_t\_r} + \alpha_5 L_{cr\_emb}
\label{loss}
\end{aligned}
\end{equation} 

During inference, for each distinct text word $w_k$ in training data, we compute its text phonetic vector $(\mathbf{v_{p_t}})_{k}$, k = 1, ..., N.
Then for each spoken word $\mathbf{x}_i$ in testing data, we apply softmax on the negative distance between its audio phonetic vector $\mathbf{v_{p_a}}$ and each text phonetic vector $(\mathbf{v_{p_t}})_{k}$ to get the posterior probability for each text word $Pr_{a}(w_k|\mathbf{x}_i)$:
\begin{equation}
\begin{aligned}
Pr_{a}(w_k|\mathbf{x}_i) &= \frac{\exp(-\| E_p(\mathbf{x}_i) - (\mathbf{v_{p_t}})_{k} \|_2^2)}{\sum_{j=1}^N \exp(-\| E_p(\mathbf{x}_i) - (\mathbf{v_{p_t}})_{j} \|_2^2)}.
\label{pr_a}
\end{aligned}
\end{equation} 

When a large amount of unpaired text data is available, a language model can be trained and integrated into the inference.
Suppose the spoken word $\mathbf{x}_i$ is the $t$-th spoken word in an utterance $\mathbf{u}$ and its corresponding text word is $u_t$.
The log probability for recognition is then:
\begin{equation}
\begin{aligned}
\log Pr(u_t=w_k|\mathbf{x}_i) &= \log Pr_{a}(w_k|\mathbf{x}_i) \\
&+ \beta \log Pr_{LM}(u_t=w_k),
\label{pr}
\end{aligned}
\end{equation} 
where the first term is as in (\ref{pr_a}), and $Pr_{LM}(\cdot)$ is the language model score.
All $\alpha_i$ and $\beta$ above are hyperparameters.
\vspace{-1mm}

\subsection{Cycle Consistency Regularization}
\label{subsec:cycle}

\vspace{-1mm}
We can further add a cycle-consistency loss to the original loss function (\ref{loss}):
\begin{equation}
\begin{aligned}
L_{cycle} &= \sum_{(\mathbf{x}_j, \mathbf{y}_j) \in \mathbf{Z}} \| \mathbf{x}_j - D_a(E_t(D_t(E_p(\mathbf{x}_j))), E_s(\mathbf{x}_j)) \|_2^2 \\
&+ \| \mathbf{y}_j - D_t(E_p(D_a(E_t(\mathbf{y}_j), E_s(\mathbf{x}_j)))) \|_2^2. 
\label{cycle_loss}
\end{aligned}
\end{equation}
Part of the first term was shown by the dotted purple cycle in the right of Figure~\ref{fig:overview}, while part of the second term was shown by the dotted blue loop in the left of the figure.
\vspace{-1mm}

 



\begin{table}[t]
\footnotesize
\centering
\caption{Word error rate (WER) (\%) performance spectrum for different training data sizes and different N (number of paired words) with joint learning in (\ref{loss}) of Subsection~\ref{subsec:inference}.}
\label{table:unpair_num}
\begin{tabular}{|c||c|c|c|c|c|c|}
\hline

\multicolumn{1}{|c||}{N (\# of} &
\multicolumn{6}{c|}{Total Speech Data Size (Paired plus unlabeled)} \\ \cline{2-7}

\multicolumn{1}{|c||}{paired)} & 
\multicolumn{1}{c|}{0.1hr} &
\multicolumn{1}{c|}{0.25hr} &
\multicolumn{1}{c|}{0.5hr} &
\multicolumn{1}{c|}{1.0hr} &
\multicolumn{1}{c|}{\textcolor{red}{2.1hr}} &
\multicolumn{1}{c|}{\textcolor{blue}{4.1hr}} \\ \hline \hline

39809 & - & - & - & - & - & 32.9 \\ \hline
\textcolor{blue}{20000} & - & - & - & - & 36.6 & \textcolor{blue}{34.2} \\ \hline
10000 & - & - & - & 42.3 & 38.4 & 36.4 \\ \hline
5000 & - & - & 48.2 & 44.8 & 41.0 & 38.9 \\ \hline
\textcolor{red}{2500} & - & 55.3 & 50.3 & 47.1 & \textcolor{red}{44.6} & 42.5 \\ \hline
1000 & 65.0 & 57.8 & 54.2 & 51.5 & 50.2 & 48.2 \\ \hline
600 & 65.2 & 61.7 & 57.9 & 56.5 & 55.4 & 54.7 \\ \hline
200 & 69.7 & 69.1 & 67.6 & 68.9 & 67.4 & 67.6 \\ \hline
100 & 77.2 & 76.3 & 78.0 & 78.3 & 82.8 & 78.7 \\ \hline
50 & 82.8 & 81.8 & 80.5 & 80.0 & 82.1 & 85.4 \\ \hline

\end{tabular}
\vspace{-3mm}
\end{table}

\subsection{Separate Learning then Transformation}
\label{subsec:dis_spk}

\vspace{-1mm}
Because the continuous signals of spoken words and the discrete symbol sequences of text words are very different, the alignment between the two latent spaces as mentioned above may not be smooth.
For example, during the joint learning in (\ref{loss}) the cross-domain losses (\ref{cr_a_r_loss})(\ref{cr_t_r_loss})(\ref{cr_emb_loss}) inevitably disturb the intra-domain losses (\ref{in_a_r_loss})(\ref{in_t_r_loss}) and produce distortions on the phonetic structures for the individual audio and text domains.
Of course there exist a different option, i.e., training the intra-domain phonetic structures separately for spoken and text words first, and then learn a transformation between them.

This concept can be understood by replacing the red dotted block in the middle right of Figure~\ref{fig:overview} denoted by ``Embedding Alignment" by that shown in Figure~\ref{fig:alignment}.
In this way Figure~\ref{fig:overview} becomes two independent autoencoders, for spoken and text words on the left and right, respectively trained with intra-domain reconstruction losses (\ref{in_a_r_loss})(\ref{in_t_r_loss}) only, plus a set of alignment transformations $\mathbf{M_{at}}$ and $\mathbf{M_{ta}}$ between the two latent spaces.
In this way the phonetic structures over the spoken and text words may be better learned separately in different spaces. In the left part of Figure~\ref{fig:overview}, however, a set of GAN-based~\cite{chen2018almost,WGANGP} criteria is needed to disentangle the speaker characteristics from phonetic structures (not shown in Figure~\ref{fig:overview}), while in the original Figure~\ref{fig:overview} this disentanglement can be achieved with the help from the text word autoencoder.

\begin{figure}[t]
  \centering
  \includegraphics[width=\linewidth]{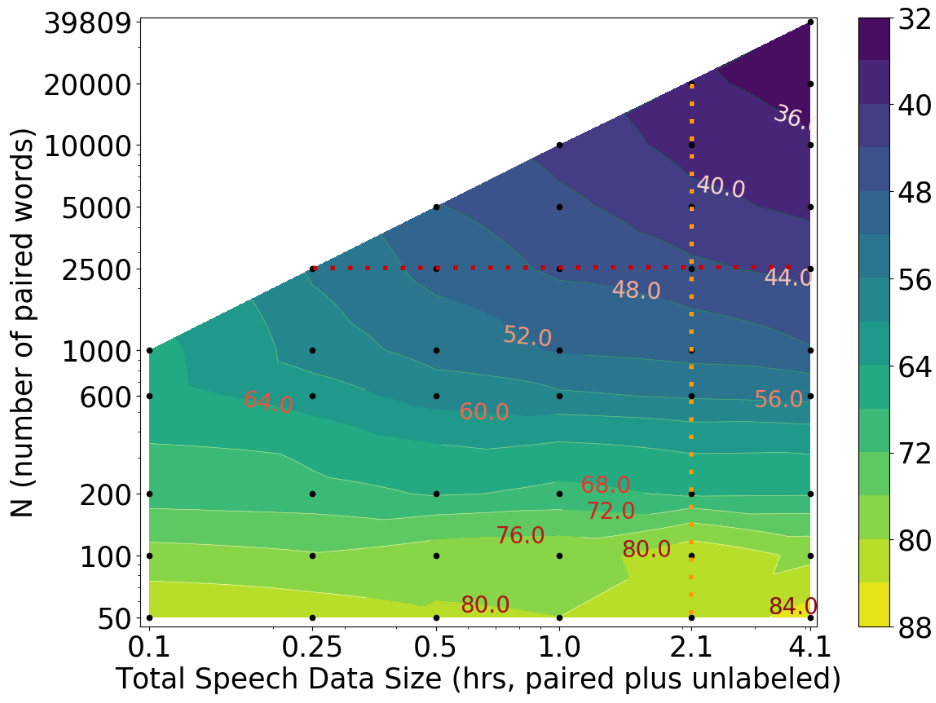}
  \caption{The 2-dim display of the WER (\%) performance spectrum of Table~\ref{table:unpair_num} for different training data sizes (hrs), and different N (number of paired words). The black dots are the real experimental results. The contours are produced based on linear interpolation among black dots.}
  \label{fig:contour}
\vspace{-2mm}
\end{figure}

The phonetic vectors $\mathbf{v_{p_a}}$ and $\mathbf{v_{p_t}}$ separately trained are first normalized in all dimensions and projected onto their lower dimensional space by PCA.
The projected vectors in the principal component spaces are respectively denoted as $\mathbf{A} = \{\mathbf{a}_i\}_{i=1}^M$ for audio and $\mathbf{T} = \{\mathbf{t}_i\}_{i=1}^N$ for text.
The paired spoken and text words, $\mathbf{Z} = \{(\mathbf{x}_j, \mathbf{y}_j)\}$, are denoted here as $\mathbf{Z} = \{(\mathbf{a}_j, \mathbf{t}_j)\}$, in which $\mathbf{a}_j$ and $\mathbf{t}_j$ correspond to the same word.
Then a pair of transformation matrices, $\mathbf{M_{at}}$ and $\mathbf{M_{ta}}$ are learned, where $\mathbf{M_{at}}$ maps a vector $\mathbf{a}$ in $\mathbf{A}$ to the space of $\mathbf{T}$, that is, $\mathbf{t} = \mathbf{M_{at}}\mathbf{a}$, while $\mathbf{M_{ta}}$ maps a vector $\mathbf{t}$ in $\mathbf{T}$ to the space of $\mathbf{A}$.
$\mathbf{M_{at}}$ and $\mathbf{M_{ta}}$ are initialized as identity matrices and then learned iteratively with gradient descent minimizing the objective function:
\begin{equation}
\begin{aligned}
L_t = \sum_{(\mathbf{a}_j, \mathbf{t}_j) \in \mathbf{Z}} \|\mathbf{t}_j - \mathbf{M_{at}}\mathbf{a}_j \|_2^2 +  \sum_{(\mathbf{a}_j, \mathbf{t}_j) \in \mathbf{Z}} \|\mathbf{a}_j - \mathbf{M_{ta}}\mathbf{t}_j \|_2^2  \\
+ \lambda^\prime \sum_{(\mathbf{a}_j, \mathbf{t}_j) \in \mathbf{Z}} (\| \mathbf{a}_j - \mathbf{M_{ta}}\mathbf{M_{at}}\mathbf{a}_j \|_2^2 + \| \mathbf{t}_j - \mathbf{M_{at}}\mathbf{M_{ta}}\mathbf{t}_j \|_2^2).
\end{aligned} \label{eq:align}
\end{equation}
In the first two terms, we want the transformation of $\mathbf{a}_j$ by $\mathbf{M_{at}}$ to be close to $\mathbf{t}_j$ and vice versa.
The last two terms are for cycle consistency, i.e., after transforming $\mathbf{a}_j$ to the space of $\mathbf{T}$ by $\mathbf{M_{at}}$ and then transforming back by $\mathbf{M_{ta}}$, it should end up with the original $\mathbf{a}_j$, and vice versa.
$\lambda^\prime$ is a hyper-parameter.
\vspace{-1mm}



\begin{table}[t]
\footnotesize
\centering
\caption{Comparison between Joint Learning of (\ref{loss}) in Subsection~\ref{subsec:inference} and Separate Learning then transformation of (\ref{eq:align}) in Subsection~\ref{subsec:dis_spk} for L=1,2,3 layers of GRUs, using phonemes or characters (abbrv. as ``Char." in the table) as the subword units, with 4.1 hrs of data and N=39809 and 1000.}
\label{table:jnt_sep}
\begin{tabular}{|c||c|c|c||c|c|c||c|}
\hline

\multicolumn{1}{|c||}{N} &
\multicolumn{6}{c||}{Phoneme as the subword unit} &
\multicolumn{1}{c|}{Char.} \\ \cline{2-8}

\multicolumn{1}{|c||}{(\# of} &
\multicolumn{3}{c||}{Joint (with (\ref{loss}))} &
\multicolumn{3}{c||}{Separate (with (\ref{eq:align}))} &
\multicolumn{1}{c|}{Joint} \\ \cline{2-8} 

\multicolumn{1}{|c||}{paired)} & 
 \multicolumn{1}{c|}{L=1} &
\multicolumn{1}{c|}{L=2} &
\multicolumn{1}{c||}{L=3} &
\multicolumn{1}{c|}{L=1} &
\multicolumn{1}{c|}{L=2} &
\multicolumn{1}{c||}{L=3} &
\multicolumn{1}{c|}{L=1} \\ \hline \hline

39809 & 32.9 & 31.7 & \textbf{31.3} & 68.6 & 67.0 & 65.6 & 38.6 \\ \hline
1000 & 48.2 & \textbf{47.4} & 47.5 & 71.8 & 72.5 & 74.6 & 60.5\\ \hline

\end{tabular}
\vspace{-5mm}
\end{table}





\begin{table}[t]
\footnotesize
\centering
\caption{Ablation studies for the proposed approach of joint learning in (\ref{loss}) when removing a loss term in (\ref{in_a_r_loss})(\ref{in_t_r_loss})(\ref{cr_a_r_loss})(\ref{cr_t_r_loss})(\ref{cr_emb_loss}) with 4.1 hrs of data.}
\label{table:ablation}
\begin{tabular}{|c|c||c|c|c|c|c|c|}
\hline

\multicolumn{2}{|c||}{Loss} &
\multicolumn{1}{c|}{(\ref{loss})} &
\multicolumn{1}{c|}{- (\ref{in_a_r_loss})} &
\multicolumn{1}{c|}{- (\ref{in_t_r_loss})} &
\multicolumn{1}{c|}{- (\ref{cr_a_r_loss})} &
\multicolumn{1}{c|}{- (\ref{cr_t_r_loss})} &
\multicolumn{1}{c|}{- (\ref{cr_emb_loss})} \\ \hline \hline

\multirow{2}{*}{N} &
39809 & \textbf{32.9} & 33.1 & 33.9 & 33.2 & 44.8 & 50.4 \\ \cline{2-8}
 & 1000 & \textbf{48.2} & 51.4 & 49.3 & 48.6 & 57.0 & 69.5 \\ \hline



\end{tabular}
\vspace{-1mm}
\end{table}

\begin{table}[t]
\footnotesize
\centering
\caption{Contributions by the cycle-consistency in (\ref{cycle_loss}) of Subsection~\ref{subsec:cycle} for 4.1 hrs of data and different N.}
\label{table:cycle}
\begin{tabular}{|c||c|c|c|c|c|}
\hline

\multirow{2}{*}{Loss} &
\multicolumn{5}{c|}{N (number of paired words)} \\ \cline{2-6}
\multicolumn{1}{|c||}{} &
\multicolumn{1}{c|}{39809} &
\multicolumn{1}{c|}{1000} &
\multicolumn{1}{c|}{200} &
\multicolumn{1}{c|}{100} &
\multicolumn{1}{c|}{50} \\ \hline \hline

(\ref{loss}) & \textbf{32.9} & \textbf{48.2} & 67.6 & 78.7 & 85.4 \\ \hline
Plus cycle (\ref{cycle_loss}) & 41.4 & 51.5 & \textbf{66.4} & \textbf{74.7} & \textbf{82.3} \\ \hline

\end{tabular}
\vspace{-4mm}
\end{table}









\section{Experimental Setup}
\label{sec:exp}
\vspace{-1mm}

\subsection{Dataset}
\label{subsec:dataset}

\vspace{-1mm}
TIMIT~\cite{garofolo1993darpa} dataset was used here. 
Its training set contains only 4620 utterances (4.1 hours) with a total of 39809 words, or 4893 distinct words. 
So this dataset is close to the ``very-low" resource setting considered here.
We followed the standard Kaldi recipe~\cite{povey2011kaldi} to extract the MFCCs of 39-dim with utterance-wise cepstral mean and variance normalization (CMVN) applied as the acoustic features.  
\vspace{-1mm}

\subsection{Model Implementation}
\label{subsec:implementation}

\vspace{-1mm}
The three encoders $E_p$, $E_s$ and $E_t$ in Figure~\ref{fig:overview} were all Bi-GRUs with hidden layer size 256.
The decoders $D_a$ and $D_t$ were GRUs with hidden layer size 512 and 256 respectively.
The value of threshold $\lambda$ in (\ref{cr_emb_loss}) was set to 0.01.
Hyperparameters $(\alpha_1, \alpha_2, \alpha_3, \alpha_4, \alpha_5, \beta)$ were set to $(0.2, 1.0, 0.2, 1.0, 5.0, 0.01)$.
We trained a tri-gram language model on the transcriptions of TIMIT data and performed beam search with beam size 10 during inference in (\ref{pr}) to obtain the recognition results.
Adam optimizer~\cite{DBLP:journals/corr/KingmaB14} was used and the initial learning rate was $10^{-4}$.
The mini-batch size was 32.
In realizing the embedding alignment in Figure~\ref{fig:alignment}, the discriminator used in the audio embedding for disentangling the speaker vector was a two-layer fully-connected network with hidden size 256, and the mapping functions $\mathbf{M_{at}}$ and $\mathbf{M_{ta}}$ were linear matrices, following the setting of the previous work~\cite{chen2018almost}.
\vspace{-1mm}

\section{Experiments}
\label{sec:exp}
\vspace{-1mm}

\subsection{Performance Spectrum for Different Training Data Sizes and Different Number of Paired Words}
\label{subsec:exp_data_amount}

\vspace{-1mm}
First we wish to see the achievable performance in word error rates (WER) (\%) over the testing set for the joint learning approach of (\ref{loss}) in Subsection~\ref{subsec:inference} when the training data size and the numbers of paired words (N) are respectively reduced to as small as possible.
All the encoders and decoders are single-layer GRUs.
The results are listed in Table~\ref{table:unpair_num} (blank for upper left corner because only smaller number of words can be labeled and made paired for smaller data size).
A 2-dim display of this performance spectrum is shown in Figure~\ref{fig:contour}, where the black dots are the real results in Table~\ref{table:unpair_num}, while the contours are produced based on linear interpolation among black dots.

We can see from Table~\ref{table:unpair_num} only 2.1 hr of total data (in which 2500 spoken words were labeled and the rest unlabeled) gave an WER of 44.6\% (in red), and this number can be reduced to 34.2\% if 4.1 hr of data (in which 20000 words labeled) were available (in blue).
We can also see how the WER varied when the total data size was changed for a fixed value of N (e.g. N=2500, the horizontal dotted red line in Figure~\ref{fig:contour}) or N was changed for a fixed data size (e.g. 2.1 hr, the vertical orange line in Figure~\ref{fig:contour}).
Right now these numbers are still relatively high (specially for $N \leq 1000$ or less than 1.0 hr of data), but the smooth, continuous performance spectrum may imply the proposed approach is a good direction and better performance may be achievable in the future.
For example, the aligned phonetic structures for the N paired words seemed to ``generalize" to more words not labeled.
Also, it can be found that in the lower half of Figure~\ref{fig:contour} the contours are more horizontal, implying for small N (e.g. $N \leq 600$) the help offered by larger data size may be limited.
In the upper half of the figure~\ref{fig:contour}, however, the contours go up on the left implying for larger N (e.g. $N \geq 600$) larger data size gave lower WER.
\vspace{-1mm}

\subsection{Different Learning Strategies and Model Parameters}
\label{subsec:exp_approaches}

\vspace{-1mm}
Table~\ref{table:unpair_num} and Figure~\ref{fig:contour} are for the joint learning strategy in (\ref{loss}) of Subsection~\ref{subsec:inference} and single-layer GRUs.
Here we wish to see the performance for the strategy of separate learning plus a transformation afterwards in (\ref{eq:align}) of Subsection~\ref{subsec:dis_spk}.
The results are in the left section (Joint) and middle section (Separate) of Table~\ref{table:jnt_sep}, for 4.1 hrs of data and N=39809, 1000.
Results for 2 and 3 layers of GRUs in encoders/decoders (L=2, 3) are also listed.

The results in Table~\ref{table:jnt_sep} empirically showed joint learning the phonetic structures from spoken and text words together with the alignment between them outperformed the strategy of separate learning then transformation.
Very probably the phonetic structures of subword unit sequences of given length are very different from those of signal segments of different length, so aligning and warping them during joint learning gives smoother mapping relationships, while a forced transformation between two separately trained structures may be too rigid.
Also, the model with L=2 achieved slightly better results in comparison with L=1 in the case of 4.1 hrs of data here, while overfitting happened with L=3 when N was small.
All the above results are for phonemes taken as the subword units.
The right column of Table~\ref{table:jnt_sep} are the results for characters instead with joint learning and L=1.
We see characters worked much worse than phonemes.
Clearly the phoneme sequences described the phonetic structures much better than character sequences.
\vspace{-1mm}

\subsection{Ablation Studies and Cycle-consistency Regularization}
\label{subsec:exp_ablation}

\vspace{-1mm}
In Table~\ref{table:ablation}, we performed ablation studies for joint learning in (\ref{loss}) of Subsection~\ref{subsec:inference} and 4.1 hrs data and N=39809 and 1000 by removing a loss term in (\ref{in_a_r_loss})(\ref{in_t_r_loss})(\ref{cr_a_r_loss})(\ref{cr_t_r_loss})(\ref{cr_emb_loss}) in Subsection~\ref{subsec:unsupervised},~\ref{subsec:cross-domain} and~\ref{subsec:emb_align}.
We see all reconstruction losses in (\ref{in_a_r_loss})(\ref{in_t_r_loss})(\ref{cr_a_r_loss})(\ref{cr_t_r_loss}) were useful, but the cross-domain text reconstruction loss in (\ref{cr_t_r_loss}) was specially important, obviously because the phoneme sequences described the phonetic structures most precisely, and the cross-domain reconstruction offered good mapping relationships.
On the other hand, the cross-domain embedding loss learning from paired spoken and text words in (\ref{cr_emb_loss}) made the most significant contribution here.
The knowledge learned here from paired data ``generalize" to other unlabeled words.

We also tested the cycle-consistency mentioned in (\ref{cycle_loss}) of Subsection~\ref{subsec:cycle} for 4.1 hrs of data and different N.
The results in Table~\ref{table:ablation} showed that the cycle consistency may not help for larger N, but became useful for smaller N (e.g. $N \leq 200$) when too few number of such paired words or ``anchor points" were inadequate for constructing the mapping relationships.
This is because the cycle-consistency condition required every paired spoken and text word to go through all encoders and decoders.
\vspace{-1mm}

\section{Discussion and Conclusion}
\label{sec:end}
\vspace{-1mm}

In this work, we investigate the possibility of performing speech recognition with very low resource (small data size with small number of paired labeled words) by joint learning the phonetic structures from audio and text embeddings.
Smooth and continuous WER performance spectrum when the data size and number of paired words were respectively reduced to as small as possible was obtained.
The achieved WERs were still relatively high, but implied a good direction for future work.
\vspace{-1mm}

\bibliographystyle{IEEEtran}
\bibliography{mybib,IR_bib,ref_dis,segment,transfer,INTERSPEECH16,ICASSP13,refs}

\end{document}